# Human Face Recognition using Gabor based Kernel Entropy Component Analysis

Arindam Kar[1], Debotosh Bhattacharjee[2], Dipak Kumar Basu[2*], Mita Nasipuri[2], Mahantapas Kundu[2]

[1] Indian Statistical Institute, Kolkata, India

[2] Department of Computer Science and Engineering, Jadavpur University, Kolkata, India

* Former Professor and AICTE Emeritus Fellow

{kgparindamkar@gmail.com, debotosh@indiatimes.com, dipakkbasu@gmail.com, m.nasipuri@cse.jdvu.ac.in, mkundu@cse.jdvu.ac.in}

**Abstract.** In this paper, we present a novel Gabor wavelet based Kernel Entropy Component Analysis (KECA) method by integrating the Gabor wavelet transformation (GWT) of facial images with the KECA method for enhanced face recognition performance. Firstly, from the Gabor wavelet transformed images the most important discriminative desirable facial features characterized by spatial frequency, spatial locality and orientation selectivity to cope with the variations due to illumination and facial expression changes were derived. After that KECA, relating to the Renyi entropy is extended to include cosine kernel function. The KECA with the cosine kernels is then applied on the extracted most important discriminating feature vectors of facial images to obtain only those real kernel ECA eigenvectors that are associated with eigenvalues having positive entropy contribution. Finally, these real KECA features are used for image classification using the $L_1$, $L_2$ distance measures; the Mahalanobis distance measure and the cosine similarity measure. The feasibility of the Gabor based KECA method with the cosine kernel has been successfully tested on both frontal and pose-angled face recognition, using datasets from the ORL, FRAV2D and the FERET database.

## 1    Introduction

Face authentication has gained considerable attention in the recent years due to the increasing need in access verification systems, surveillance, security monitoring and so on. Such systems are used for the verification of a user's identity on the internet, bank automaton, authentication for entry to secured buildings, etc. Face recognition involves recognition of personal identity based on geometric or statistical features derived from face images. Even though humans can identify faces with ease, but developing an automated system that accomplishes such objectives is very challenging. The challenges are even more intense when there are large variations in illumination conditions, viewing directions or poses, facial expression, aging, etc.

Robust face recognition schemes require features with low-dimensional representation for storage purposes and enhanced discrimination abilities for subsequent image retrieval. The representation methods usually start with a dimensionality reduction procedure as the high-dimensionality of the original visual space makes the statistical estimation very difficult, if not impossible, due to the fact that the high-dimensional space is mostly empty. Data transformation is of fundamental importance in pattern analysis and machine intelligence. The goal is to transform the potentially high dimensional data into an alternative and typically lower dimensional representation, which reveals the underlying structure of the data. The most well known method for this is principal component analysis (PCA) [1, 2], which is based on the data correlation matrix. It is a linear method ensuring that the transformed data are uncorrelated and preserve maximally the second order statistics of the original data. This method is based on selecting the top or bottom eigenvalues (spectrum) and eigenvectors of specially constructed data matrices. Another linear method is metric multidimensional scaling (MDS), which preserves inner products. Metrics MDS and PCA can be shown to be equivalent [3,4]. In recent years, a number of advanced nonlinear spectral data transformation methods have been proposed. A very influential method is kernel PCA (KPCA) [5]. KPCA performs traditional PCA in a so-called kernel feature space, which is nonlinearly related to the input space [6]. This is enabled by a positive semi definite (psd) kernel function, which computes inner products in the kernel feature space. An inner product matrix, or kernel matrix, can thus be constructed. Performing metric MDS on the kernel matrix, based on the top eigenvalues of the matrix, provides the KPCA data transformation. KPCA has been used in many contexts. A spectral clustering method was, for example, developed in [7], where C-means [8] is performed on the KPCA eigenvectors. KPCA has also been used for pattern de-noising [9, 10, 11, 12] and classification [13]. Many other spectral methods exist, which differ in the way the data matrices are constructed. In [13], for example, the data transformation is based on the normalized Laplacian matrix, and clustering is performed using C-means on unit norm spectral data. Manifold learning and other clustering variants using the Laplacian matrix exists [14].

Kernel Entropy Component Analysis (KECA) [15] is a new and only spectral data transformation method on information theory. It is directly related to the Renyi entropy of the input space data set via a kernel-based Renyi entropy estimator. Renyi entropy estimator is expressed in terms of projections onto the principal axes of feature space. The KECA transformation is based on the most entropy preserving axes. The present work in this paper is possibly the first application of Gabor based KECA in a face recognition or image classification problem.

This paper presents a novel Gabor based kernel Entropy Component Analysis (KECA) method by integrating the Gabor wavelet transformation (GWT) of face images and the KECA for face recognition. Gabor wavelet transformation [16, 17, 18], first derives desirable facial features characterized by spatial frequency, spatial locality and orientation selectivity to cope with the variations due to illumination and facial expression changes. The most discriminating features are extracted from these Gabor transformed images, most. Finally, KECA, with cosine kernel function [19] is applied on the extracted feature vectors to enhance the face recognition performance.

The remainder of the paper is organized as follows. Section 2 describes the derivation of most discriminating feature vectors from the Gabor convolution outputs of the facial images. Section 3 details the novel Gabor based KECA method with cosine kernel function for face recognition, section 4 reports the performance of the proposed method on the face recognition task by applying it on the FERET [20], FRAV2D [21], and ORL [22] face databases and also the comparison of the proposed method with some of the most popular face recognition schemes. Finally, our paper is concluded in section 5.

## 2 Discriminating Feature extraction from Gabor Convolution Outputs

Gabor wavelet representation of face images selectively derives desirable features characterized by spatial frequency, spatial locality and orientation. These features are robust to variations due to illumination and facial expression changes [23, 24, 25, 26]. The dimensionality of Gabor representation is very high, which makes statistical estimation very difficult. To avoid such problem only the most discriminative features containing most of the information of the Gabor transformed image are extracted.

### 2.1 Gabor Wavelets Convolution Outputs

Gabor wavelets are used for image analysis because of their biological relevance and computational properties [27, 28, 29, 30]. The Gabor wavelets, whose kernels are similar to the 2-D receptive field profiles of the mammalian cortical simple cells, exhibit strong characteristics of spatial locality and orientation selectivity and are optimally localized in the space and frequency domains. The Gabor wavelets can be defined as follows [31, 32]:

$$\varphi_{\mu,\nu}(z) = \frac{\left\| k_{\mu,\nu} \right\|^2}{\sigma^2} e^{-\left\| k_{\mu,\nu} \right\|^2 \left\| z \right\|^2 / 2\sigma^2} \left[ e^{i k_{\mu,\nu}(z)} - e^{-\left( \sigma^2 / 2 \right)} \right] \qquad (1)$$

where $\mu$ and $\nu$ define the orientation and scale of the Gabor kernels, z =(x,y), and $\|.\|$ denotes the norm operator, and $k_{\mu,\nu}$ is the wave vector.

$k_{\mu,\nu}$ can be defined as: $k_{\mu,\nu} = k_{\nu} e^{i\phi_{\mu}}$, where, $k_{\nu} = k_{max} / f^{\nu}$ and $k_{max} = \pi / 2$, $\phi_{\mu} = \pi\mu / 8$, $\mu = 0,....7$, where $f$ is the spacing factor between kernels in the frequency domain. The Gabor kernels in (1) are all self-similar since they can be generated from only one filter, the mother wavelet, by scaling and rotation via the wave vector. Each kernel is a product of a Gaussian envelope and a complex plane wave, while the first term within the square brackets in equation (1) determines the oscillatory part of the kernel and the second term compensates for the dc value. The term $e^{-\left( \sigma^2 / 2 \right)}$ is subtracted from equation (1) in order to make the kernel DC-free [16]. Thus the transformed images become insensitive to illumination.

## 2.2 Gabor Feature representation

The Gabor wavelet representation of an image is the convolution of the image with a family of Gabor kernels as defined in (1). Let $I(x, y)$ be the gray level distribution of an image. Then the convolution output $(O_{\mu,\upsilon}(z))$ of an image $I$ and a Gabor kernel $\varphi_{\mu,\upsilon}$ is defined as:

$$O_{\mu,\upsilon}(z) = I(z) * \varphi_{\mu,\upsilon}(z),\qquad(2)$$

where z = (x, y), * denotes the convolution operator. Applying the convolution theorem, we can derive the convolution output from (2) via the Discrete Fourier Transform as:

$$\Im\{O_{\mu,\upsilon}(z)\} = \Im\{I(z)\}\Im\{\varphi_{\mu,\upsilon}(z)\},\qquad(3)$$

$$\text{and } O_{\mu,\upsilon}(z) = \Im^{-1}\{\Im\{I(z)\}\Im\{\varphi_{\mu,\upsilon}(z)\}\}.\qquad(4)$$

where $\Im$ and $\Im^{-1}$ denote the Discrete Fourier and inverse Discrete Fourier transform, respectively. The outputs $O_{\mu,\upsilon}(z)$ ($\upsilon \in \{0,...., 4\}$, and $\mu \in \{0,..., 7\}$) consist of different local, scale, and orientation features. The outputs exhibit strong characteristics of spatial locality, scale, and orientation which are needed for selection of salient local features, such as the eyes, nose and mouth, that are suitable for visual event recognition and thus makes them a suitable choice for facial feature extraction to derive local and discriminating features for classification of faces. The magnitude of $O_{\mu,\upsilon}(z)$ is defined as modulus of $O_{\mu,\upsilon}(z)$, i.e., $\|O_{\mu,\upsilon}(z)\|$.

## 2.3 Most Discriminating Gabor Feature Vector Representation

It is to be noted that we have considered the magnitude of $O_{\mu,\upsilon}(z)$ only excluding its phase, since it is consistent with the application of Gabor representations [33]. As the outputs $(O_{\mu,\upsilon}(\xi): \mu \in \{0,...., 4\}, \upsilon \in \{0,.... 7\})$ which, consists of 40 local scale and orientation features, the dimensionality of the Gabor feature vector space becomes very high. The method for extraction of the most discriminative features from the convolution outputs is explained in Algorithm1.

Algorithm1:

- Step1: Find the convolution outputs of the original image with all the Gabor kernels. As the convolution outputs contain complex values, so replace each pixel value of the convolution output by its modulus and the resulting image is termed as $G_I$, I=1,2,…,K; K= total no. of Gabor kernels.
- Step 2: Divide each convolution output image $G_I$ into a number of blocks each of size $L \times L$ pixels; L<M, and L<N. So the total number of blocks obtained from each convolution output is $\left\lfloor \frac{M}{L} \right\rfloor \times \left\lfloor \frac{N}{L} \right\rfloor$.

- Step 3: From each block B consider only those pixels whose values are greater than overall mean of the image and store it in a column vector, which is the most discriminating feature vector, say $\chi_i$ .

- Step 4: Repeat Step 1 to Step 3 for all the convolution outputs to obtain, a total set of $\chi_1, \chi_2, ..., \chi_E$ column vectors each of size $\left\lfloor \frac{M}{L} \right\rfloor \times \left\lfloor \frac{N}{L} \right\rfloor$ .

- Step 5: Derivation of the most discriminating feature is as follows :

  For each image $G_I^{(\alpha)}(x, y)$, $\alpha = 1, 2, ... E$ ,

  Find $\overline{g}_\alpha = \frac{1}{M \times N} \sum_{x, y} G_I^{(\alpha)}(x, y)$ (5), where $\overline{g}_\alpha$ is the mean of the overall image.

Then for each block B find its maximum value, say, $m_B$, if $(m_B) \geq \overline{g}_\alpha$ and take feature point $= m_B$; else feature point $= \overline{g}_\alpha$ . Store this feature points into the column vector $\chi_i$ .

- Step 6: Finally concatenate each of these feature vectors to form the final feature vector $\chi = [\chi_1, \chi_2, ..., \chi_E]$, which contains the most discriminating features of size $E * \left\lfloor \frac{M}{L} \right\rfloor \times \left\lfloor \frac{N}{L} \right\rfloor$, from all the Gabor convolution outputs.

This extracted feature vector $\chi$ encompasses the most informative elements of the Gabor transformed image. The block size L × L is one of the important parameter of the proposed technique, as it must be chosen small enough to capture the important discriminative features and large enough to avoid redundancy. Step 5 is applied in order not to get stuck on a local maximum. In the experiments different block sizes are used. But a 7×7 block is finally used to extract the most informative features from the 40 Gabor convolution outputs of the input image as it achieves the best results. Thus the extracted facial features can be compared locally, instead of using a general structure, allowing us to make a decision from the parts of the face.

## 3    Kernel Entropy Component Analysis (KECA)

Applying the PCA technique to face recognition, Turk and Pentland [34] developed a well known Eigenfaces method, where the eigenfaces correspond to the eigenvectors associated with the largest eigenvalues of the face covariance matrix. The eigenfaces thus define a feature space, or "face space", which drastically reduces the dimensionality of the original space, and face detection and recognition are then carried out in the reduced space. Based on PCA, a lot of face recognition methods have been developed to improve classification accuracy and generalization performance [35, 36, 37, 38]. The PCA technique, however, encodes only for second order statistics, namely the variances and the covariance's. As these second order statistics provide only partial information on the statistics of both natural images and human faces, it might become necessary to incorporate higher order statistics as well. PCA is thus extended to a nonlinear form by mapping nonlinearly the input space to a feature space, where PCA is ultimately implemented. Due to the nonlinear mapping between the input space and the feature space, this form of PCA is nonlinear and naturally called nonlinear PCA [39]. Applying different mappings, nonlinear PCA can encode arbitrary higher-order correlations among the input variables. The nonlinear mapping between the input space and the feature space, with a possibly prohibitive computational cost, is implemented explicitly by kernel PCA [40]. The KECA data transformation method is fundamentally different from other spectral methods in two very important ways as follows: a) the data transformation reveals structure related to the Renyi entropy of the input space data set and b) this method does not necessarily use the top eigenvalues and eigenvectors of the kernel matrix. This new method is called as kernel entropy component analysis or simply as KECA. In order to develop KECA, an estimator of the Renyi entropy may be expressed in terms of the spectral properties of a kernel matrix induced by a Parzen window [41] for density estimation. Further, KECA may be considered as a reasonable alternative to KPCA for pattern de-noising. The KPCA performs dimensionality reduction by selecting m eigenvalues and corresponding eigenvectors solely based on the size of the eigenvalues, and the resulting transformation may be based on uninformative eigenvectors from an entropy perspective. But Kernel entropy component analysis may be defined as a k-dimensional data transformation obtained by projecting Φ onto a subspace U$_k$ spanned by those k kernel PCA axes contributing most to the Renyi entropy estimate of the data, where Φ is a non linear mapping between the input space and the feature space. Hence, Uk is composed of a subset of the kernel PCA axes, but not necessarily those corresponding to the top k eigenvalues. This is the only spectral method which is directly related to the Renyi entropy of the input space data set via a kernel-based Renyi entropy estimator that is expressed in terms of projections onto kernel feature space principal axes. Kernel entropy component analysis is thus directly related to the Renyi entropy of the input space data set via a kernel-based Renyi entropy estimator that is expressed in terms of projections onto kernel feature space principal axes. It is to be noted that the Gaussian kernels, are the most widely used kernel functions in KECA. But in the present work cosine kernel function has been used, which further enhances face recognition as is depicted by the experimental results in section 4.4. As per the literature we have surveyed, KECA with cosine kernel function has never been used for the purpose of image classification, together with the Gabor wavelet transformation. The Renyi entropy estimator can be expressed as [41, 42]:

$$H(p) = -\log \int p^2(x)dx \quad , \quad (6)$$

where, $p(x)$ is the probability density function. As logarithm is a monotonic function, equation (6) can be formulated as $V(p) = \int p^2(x)dx = \varepsilon_p(p)$; where $\varepsilon_p(.)$, denotes expectation with respect to $x$. So, $H(p)$ can be estimated from $V(p)$ using a parzen window estimator, as [43,44]:

$$\hat{V}(p) = \frac{1}{N^2} \sum K(\chi, \chi_t) \quad , \hat{v}(p) \text{ is the estimate of } V(p). \quad (7)$$

where, $K$ is the so called parzen window or kernel defined as $K = K(\chi, \chi_t)$ centered at $\chi_t$ which computes an inner product in the Hilbert space F as:

$K\left(\chi, \chi_t\right) = \Phi\left(\chi\right)^t \cdot \Phi\left(\chi_t\right)$, where $\Phi$ is a non linear mapping between the input space and the feature space defined as: $\Phi : \mathbb{R}^N \rightarrow F$. Renyi estimator can thus be expressed in terms of the eigen values and vectors of the kernel matrix, as:

$$\hat{V}\left(p\right) = \frac{1}{N^2} \sum_{i=1}^{N} \left(\sqrt{\lambda_i} e_i^T \mathbf{1}\right)^2 \quad, \tag{8}$$

Where $\underline{\mathbf{1}}$ is a $N \times 1$ vector with all elements equal to unity. The Renyi entropy estimator can further be eigen decomposed as $K = EDE^T$, where D is diagonal matrix storing the eigenvalues $\left(\lambda_1, \lambda_2, \ldots \lambda_N\right)$ and E is a matrix with the corresponding eigen vector $e_1, \ldots e_N$ as column and the matrix K is the familiar kernel matrix as used in KPCA. Each term in this expression will contribute to the entropy estimate. Thus using KPCA the projection $\Phi$ onto the i<sup>th</sup> principal axes $u_i$ is defined as:

$$P_{u_i} \Phi = \sqrt{\lambda_i} e_i \quad, \tag{9}$$

Equation (9) therefore reveals that the Renyi entropy estimator is composed of projections onto all the kernel PCA axes. However, only a principal axis $u_i$ for which $\lambda_i > 0$ and $e_i^T \mathbf{1} \neq 0$ contributes to the entropy estimate. Those principal axes contributing most to the Renyi entropy estimate clearly carry most of the information regarding the shape of the probability density function generating the input space data set. Let $\chi$ be the extracted most discriminating Gabor feature vector, whose image in the feature space be $\Phi\left(\chi\right)$. The kernel ECA feature of $\chi$ is derived as:

$$E = P_{u_i} \Phi\left(\chi\right), \tag{10}$$

In order to derive real KECA features, we consider only those KECA eigenvectors that are associated with nonzero positive eigenvalues and sum of the components of the corresponding eigenvectors not equals to zero.

### 3.1 Kernel ECA using Cosine kernels

Let $\chi_1, \chi_2, \ldots, \chi_n \in \mathbb{R}^N$ be the data in the input space, and $\Phi$ be a nonlinear mapping between the input space and the feature space, $\Phi: R^N \longrightarrow F$. Generally three classes of kernel functions are used for nonlinear mapping: a) the polynomial kernels, b) the Radial Basis Function (RBF) kernels, c) the sigmoid kernels [45].

$$K\left(x, y\right) = \frac{\pi}{4} \cos\left(\frac{\pi\left(x \cdot y\right)}{2}\right), \tag{11}$$

The KECA produces a transformed data set with a distinct angular structure, in the sense that even the nonlinearly related input space data are distributed in different angular directions with respect to the origin of the kernel feature space. It also reveals cluster structure and reveals information about the underlying labels of the data.

These obtained cluster results are used in image classification. Classification results obtained are comparable or even better in some cases than KPCA.

### 3.2 Similarity and Classification rule

The proposed method integrates the Gabor wavelet transformation of face images together with kernel ECA with the extension of cosine kernel function models for better performance. When a face image is presented to the Gabor based KECA classifier, the low-dimensional important discriminative Gabor feature vector of the image is first calculated as detailed in Section 2, and the low-dimensional most important discriminative Gabor feature vector is used as the input data instead of the whole image in KECA to derive the KECA features, $\kappa$, using (10). Let $M_k^{'}$ be the mean of the training samples for class $w_k$, where $k = 1, 2, ..., l$ and $l$ is the number of classes. The classifier then applies the nearest neighbor (to the mean) rule for classification using the similarity (distance) measure $\delta$ :

$$\delta\left(\kappa, M_j^{'}\right) = \min_k \left(\delta\left(\kappa, M_k^{'}\right)\right) \Rightarrow \kappa \in w_j \qquad (12)$$

The Gabor based KECA feature vector, $\kappa$ is classified as belonging to the class of the closest mean, $M_k^{'}$, using the similarity measure $\delta$. The similarity measures used here are $L_1$ distance measure [46], $\delta_{L_1}$, $L_2$ distance measure[46], $\delta_{L_2}$, the Mahalanobis distance measure [47], $\delta_{Md}$, and the cosine similarity measure, $\delta_{\cos}$ [46], which are defined as:

$$\delta_{L_1} = \sum_i \left| X_i - Y_i \right|, \qquad (13)$$

$$\delta_{L_2} = \left( X - Y \right)^T \left( X - Y \right), \qquad (14)$$

$$\delta_{Md} = \left( X - Y \right)^T \Sigma^{-1} \left( X - Y \right), \qquad (15)$$

$$\delta_{\cos} = \frac{- X^T Y}{\| X \| \| Y \|}, \qquad (16)$$

Where T is the transpose operator, $\sum$ is the covariance matrix and $\| . \|$ denotes the norm operator. Note that the cosine similarity measure includes a minus sign in (16) because the nearest neighbour (to the mean) rule (12) applies minimum (distance) measure rather than maximum.

## 4 Experiments of the proposed KECA on Frontal and Pose angled Images for Face recognition

This section assesses the performance of the Gabor based KECA method for both frontal and pose angled face recognitions. The effectiveness of the Gabor based

KECA method has been tested on the dataset taken from the ORL, FRAV2D, and FERET face databases. For frontal face recognition, the data set is taken from the FRAV2D database, which contains 1100 frontal face images corresponding to 100 individuals and the whole ORL database. The images are acquired, with partially occluded face features and changes in facial expressions. For pose angled face recognition, the data set is taken from the FERET database, which contains 2200 images with different facial expressions and poses of 200 individuals. Comparative performance of the proposed method is shown against the PCA method, the LDA and the kernel PCA method. Performance results on FERET database has shown that proposed method is better than the PCA, LDA, kernel PCA and the other well known algorithms. The performance of the proposed method for both frontal and pose-angled face recognition is shown in subsections of section 4.

### 4.1 ORL Database

The whole ORL database is considered here. In the experiment, each image of this database is scaled to $92 \times 112$ pixel size with 256 gray levels. Fig. 1 shows all samples of one individual from the ORL database. The images in the ORL database have very little change in pose angle, illumination and facial expressions. The details of the images used here in the experiment are not provided here.

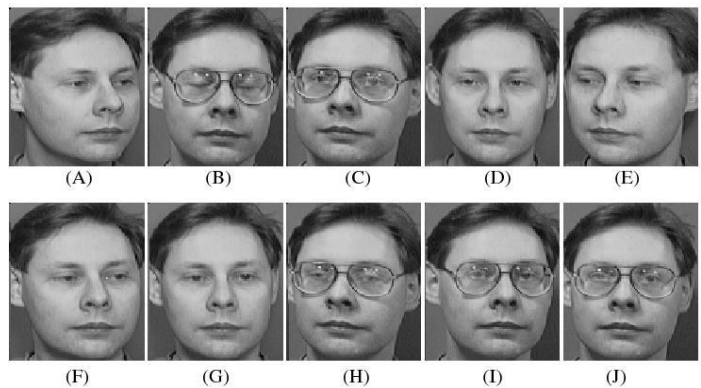

**Fig. 1**. Demonstration images of one subject from the ORL Database

### 4.2 Sensitivity and Specificity measures on ORL Database

To evaluate the sensitivity and specificity measures [48], of the proposed recognition system the dataset from the ORL database is prepared in the following manner. A total of 40 classes are formed, from the dataset of 400 images of 40 individuals. Each of the formed classes consists of 15 images, of which 10 images are of a particular individual, and rest 5 images are randomly taken from the images of other individual using permutation as shown in Fig. 2. From the 10 images of the particular individual, the first 2 images (A-B), of a particular individual is selected as the training samples and the remaining images of the particular individual are used as positive testing samples. The negative testing is done using the images of the other individual. The recognition performances in terms of true positives ($T_P$), false positive ($F_P$), true negative ($T_N$) and false negative ($F_N$); are measured. The same experiment is repeated by selecting only the first image (A) as training sample.

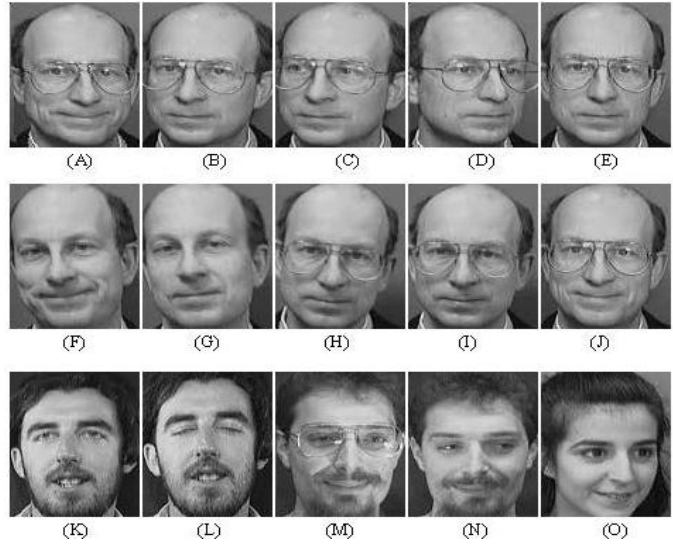

**Fig. 2.** Demonstration images of one class from the ORL database.

**Table 1: Specificity & Sensitivity measures of the ORL database**:

|  |  | Total no. of classes = 40, Total no. of images= 600 | |
|---|---|---|---|
|  |  | **Using first 2 images of an individual as training images** | |
|  |  | **Positive** | **Negative** |
| **ORL test** | **Positive** | $T_P$= 360 | $F_P$ = 0 |
|  | **Negative** | $F_N$ =0 | $T_N$ = 200 |
|  |  | Sensitivity=$T_P$/($T_P$+$F_N$)=100% | Specificity=$T_N$/($F_P$+$T_N$)=100% |
|  |  | **Using only the 1st image of an individual as training image** | |
|  |  | **Positive** | **Negative** |
| **ORL test** | **Positive** | $T_P$= 360 | $F_P$ = 0 |
|  | **Negative** | $F_N$ =0 | $T_N$ = 200 |
|  |  | Sensitivity=$T_P$/($T_P$+$F_N$)=100% | Specificity=$T_N$/($F_P$+$T_N$)=100% |

Taking the **first 2 images**, i.e., **(A-B)** of an individual for training the achieved rates are:

False positive rate = $F_P$ / ($F_P$ + $T_N$) = 1 − Specificity =0%
False negative rate = $F_N$ / ($T_P$ + $F_N$) = 1 − Sensitivity=0%
Recognition Accuracy = ($T_P$+$T_N$)/($T_P$+$T_N$+$F_P$+$F_N$) =100%.

Then, considering only the **first image (A)** of a particular individual for training the achieved rates are:

False positive rate = $F_P$ / ($F_P$ + $T_N$) = 1 − Specificity =0%
False negative rate = $F_N$ / ($T_P$ + $F_N$) = 1 − Sensitivity=0%
Recognition Accuracy = ($T_P$+$T_N$)/($T_P$+$T_N$+$F_P$+$F_N$) =100%.

The result obtained by applying the proposed approach on the ORL database outperforms, most of the previous methods are shown in table 2.

**Table 2: Comparison of Recognition performance of the proposed method with some other methods on the ORL database. The best result is shown in bold font.**

| Methods | Sensitivity Rate (%) | Reference |
|---|---|---|
| Local Gabor wavelet, PCA,ICA | 66,82.8,85 | Huang et al. (2004) [49] |
| Eigenface, Fisherface | 72.1, 76.3 | Yin et. al. (2005) [50] |
| Elastic matching | 80 | Zhang et.al.(1997) [51] |
| Gabor Filters + rank correlation | 95 | O. Ayinde and Y. Yang (2002) [52] |
| **Proposed Gabor based KECA Method** | **100** | **This paper** |

### 4.3    FERET Database

The FERET database, employed in the experiment here contains, 2,200 facial images corresponding to 200 individuals with each individual contributing 11 images. The images in this database were captured under various illuminations and display a variety of facial expressions and poses. As the images include the background and the body chest region, so each image is cropped to exclude those and are transformed into gray images and finally scaled to 92×112 pixel size. Fig. 3 shows all samples of one subject. The details of the images are as follows: (A) regular facial status; (B) +15˚pose angle; (C) -15˚ pose angle; (D) +25˚pose angle; (E) -25˚pose angle; (F) +40˚pose angle; (G) -40˚pose angle; (H) +60˚pose angle; (I) -60˚pose angle; (J) alternative expressions; (K) different illumination conditions.

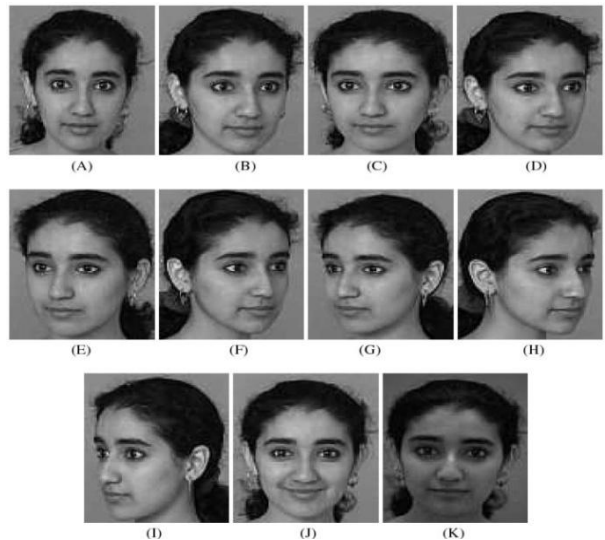

**Fig 3**. Demonstration images of an individual from the FERET Database

**Table 3**. **Average recognition results using FERET database**:

| Method | Recognition Rates(%) No. of training samples | | Avg. Recognition Rates(%) |
|---|---|---|---|
| | **3** | **4** | |
| GWT | 79 | 83.5 | 81.25 |
| **KECA** | 70 | 79 | 74.5 |
| **Gabor based KECA (with Gaussian Kernel)** | 91.5 | 94.5 | 93 |
| **Gabor based KECA (with Cosine Kernel)** | **92.6** | **96.5** | **94.5** |

### 4.4 Sensitivity and Specificity measures on the FERET dataset

To measure the sensitivity and specificity, the dataset from the FERET database is prepared in the following manner. For each individual a single class is constituted with 18 images. Thus a total 200 class is obtained, from the dataset of 2200 images of 200 individuals. Out of these 18 images in each class, 11 images are of a particular individual, and 7 images are of other individuals taken using permutation as shown in Fig. 4. Using this dataset the recognition performance in terms of the true positive ($T_P$); false positive ($F_P$); true negative ($T_N$); false negative ($F_N$); are measured. From the 11 images of the particular individual, the first 4 images (A-D), of a particular individual are selected as training samples and the remaining images of that particular individual are used as positive testing samples. The negative testing is done using the images of the other individuals. Fig. 4 shows all sample images of one class of the data set used from FERET database. Next the same process is repeated by taking only the first three images (A-C) as training samples.

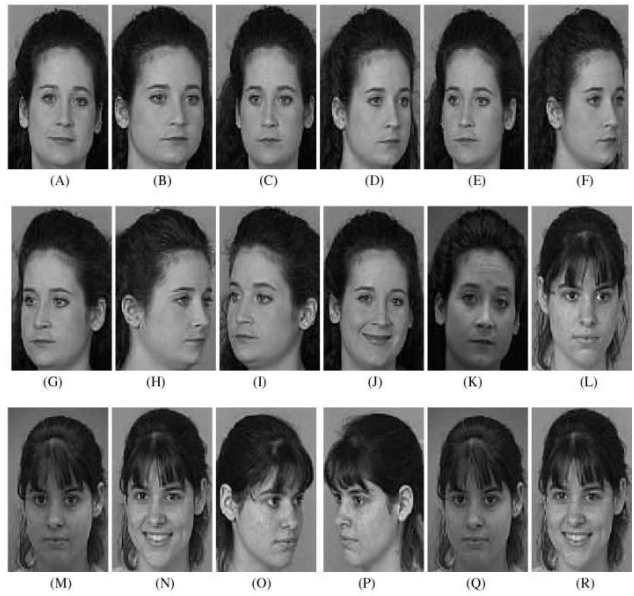

**Fig. 4**.Demonstration images of one class from the FERET database

**Table 4**: Specificity & Sensitivity measures on the FERET database:

| | | **Total no. of classes = 200, Total no. of images= 3600** | |
|---|---|---|---|
| | | **Using first 3 images of an individual as training images** | |
| | | **Positive** | **Negative** |
| **FERET test** | **Positive** | $T_P$=1481 | $F_P$ = 31 |
| | **Negative** | $F_N$ = 119 | $T_N$ = 1369 |
| | | Sensitivity=$T_P$/($T_P$+$F_N$)≈92.6% | Specificity=$T_N$/($F_P$+$T_N$)≈97.78% |
| | | **Using first 4 images of an individual as training images** | |
| | | **Positive** | **Negative** |
| **FERET test** | **Positive** | $T_P$ =1351 | $F_P$=20 |
| | **Negative** | $F_N$ = 49 | $T_N$=1380 |
| | | Sensitivity=$T_P$/($T_P$+$F_N$)=96.5% | Specificity=$T_N$/($F_P$+$T_N$)≈98.57% |

Thus, considering **first 4 images (A-D)** of an individual for training the achieved rates are:

False positive rate = $F_P$ / ($F_P$ + $T_N$) = 1 − Specificity =1.43%
False negative rate = $F_N$ / ($T_P$ + $F_N$) = 1 − Sensitivity=6.0%
Recognition Accuracy = ($T_P$+$T_N$)/($T_P$+$T_N$+$F_P$+$F_N$) ≈96.3%.

Next, considering **first 3 images (A-C)** of an individual for training the achieved rates are:

False positive rate = $F_P$ / ($F_P$ + $T_N$) = 1 − Specificity =2.28%
False negative rate = $F_N$ / ($T_P$ + $F_N$) = 1 − Sensitivity=11.5%
Recognition Accuracy = ($T_P$+$T_N$)/($T_P$+$T_N$+$F_P$+$F_N$) ≈93.1%.

## 4.5 FRAV2D Database

The FRAV2D database, employed in the experiment consists of 1100 colour face images of 100 individuals. 11 images of each individual are taken, including frontal views of faces with different facial expressions and under different illumination variations. All colour images are transformed into gray images and scaled to 92×112 pixel size. The details of the images are as follows: (A) regular facial status; (B) and (C) are images with a 15˚ turn with respect to the camera axis; (D) and (E) are images with a 30˚ turn with respect to the camera axis.(F) and (G) are images with gestures, such as smiles, open mouth, winks, laughs;(H) and (I) are images with occluded face features; (J) and (K) are images with change of illumination. Fig. 5 shows all samples of one individual.

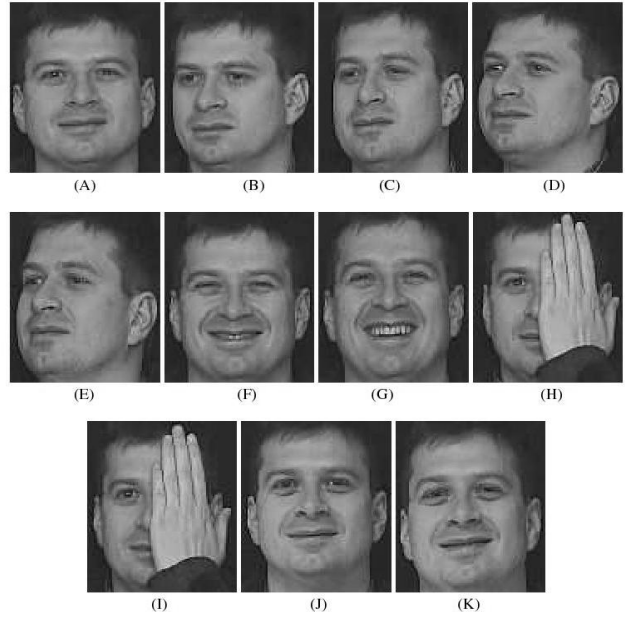

**Fig 5.** Demonstration images of one subject from the FRAV2D database

**Table 5**. **Average recognition results using FRAV2D database:**

| Method | Recognition Rates(%) No. of training samples | | Avg. Recognition Rates (%) |
|---|---|---|---|
| | **3** | **4** | |
| GWT | 84 | 87.5 | 85.75 |
| KECA | 75 | 83.25 | 79.4 |
| Gabor based KECA (with Gaussian Kernel) | 93.75 | 95.5 | 94.625 |
| **Gabor based KECA (with Cosine Kernel)** | **96.6** | **97.5** | **97.05** |

### 4.6 Sensitivity and Specificity measures on the FRAV2D dataset

To measure the sensitivity and specificity of the recognition system on FRAV2D database the dataset is prepared in the following manner. A total of 100 classes is obtained, from the dataset of 1100 images of 100 individuals. Each of these newly created classes contains 18 images, out of which 11 images are of that particular individual, and 7 images are of other individuals taken using permutation as shown in Fig. 6. Using this dataset the true positive ($T_P$); false positive ($F_P$); true negative ($T_N$) and false negative ($F_N$) are measured. From the 11 images of the particular individual, the first 4 images (A-D), of a particular individual are selected as training samples and the remaining images of that particular individual are used as positive testing samples. Negative testing is done using the images of the other individuals. Fig.6 shows all sample images of one class of the data set used from FRAV2D database. Next first three images (A-C) are used as training samples and the same process is repeated.

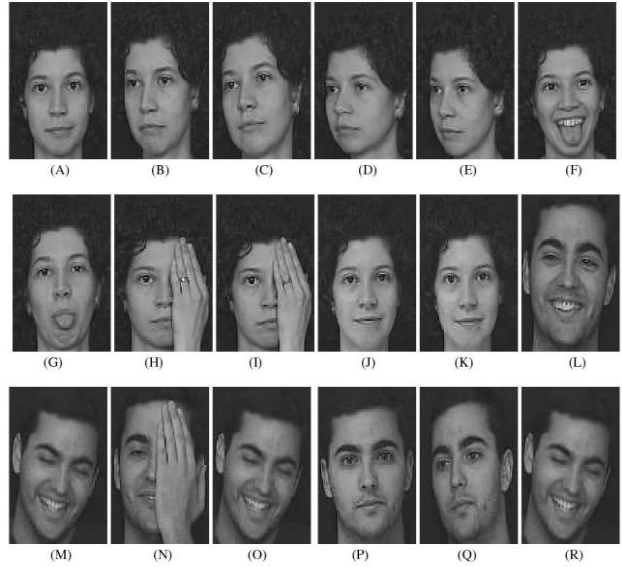

**Fig 6**. Demonstration images of one class from the FRAV2D database

**Table 6**: **Specificity & Sensitivity measure of the FRAV2D database:**

| | | Total  no. of classes = 100, Total no. of images= 1800 | |
|---|---|---|---|
| | | Using first 3 images  of an individual as training images | |
| | | **Positive** | **Negative** |
| **FRAV2D test** | **Positive** | $T_P$ = 773 | $F_P$ = 7 |
| | **Negative** | $F_N$=27 | $T_N$ = 693 |
| | | Sensitivity=$T_P/(T_P+F_N)$≈96.6% | Specificity=$T_N/(F_P+T_N)$=99.0% |
| | | Using first 4 images  of an individual as training images | |
| | | **Positive** | **Negative** |
| **FRAV2D test** | **Positive** | $T_P$ = 780 | $F_P$ = 2 |
| | **Negative** | $F_N$=20 | $T_N$ = 698 |
| | | Sensitivity=$T_P/(T_P+F_N)$≈97.5% | Specificity=$T_N/(F_P+T_N)$≈99.7% |

Taking, the **first 4 images (A-D)** of individuals for training the achieved rates are:

False positive rate = $F_P$ / ($F_P$ + $T_N$) = 1 − Specificity =.3%

False negative rate = $F_N$ / ($T_P$ + $F_N$) = 1 − Sensitivity=2.5%

Accuracy = ($T_P$+$T_N$)/($T_P$+$T_N$+$F_P$+$F_N$) ≈98.6%.

Taking the **first 3 images (A-C)** of individuals for training the achieved rates are:

False positive rate = $F_P$ / ($F_P$ + $T_N$) = 1 − Specificity =1%

False negative rate = $F_N$ / ($T_P$ + $F_N$) = 1 − Sensitivity=3.4%

Accuracy = ($T_P$+$T_N$)/($T_P$+$T_N$+$F_P$+$F_N$) ≈97.8%.

## 4.7 Experimental Results

Experiments were conducted that implements Gabor based KECA method, with four different similarity measures on both the FERET, ORL and FRAV2D database to measure both the positive and negative recognition rates i.e., the sensitivity and specificity. Fig. 7 and Fig. 8 show the face recognition performance of the proposed method in terms of specificity and sensitivity, respectively using the four different similarity measures. From both the figures it is seen that the similarity measure using Mahalanobis distance performs the best, followed in order by the $L_2$ distance measure, the $L_1$ distance measure, and the cosine similarity measure. The reason for such an ordering is that the Mahalanobis distance measure counteracts the fact that $L_2$ and $L_1$ distance measures in the PCA space weight preferentially for low frequencies. As the $L_2$ distance measure weights more the low frequencies than $L_1$ does, the $L_2$ distance measure perform better than the $L_1$ distance measure, as validated by our experiments. Fig. 9 and Fig. 10 shows the specificity measures using only the Mahalanobis distance as it performs the best.

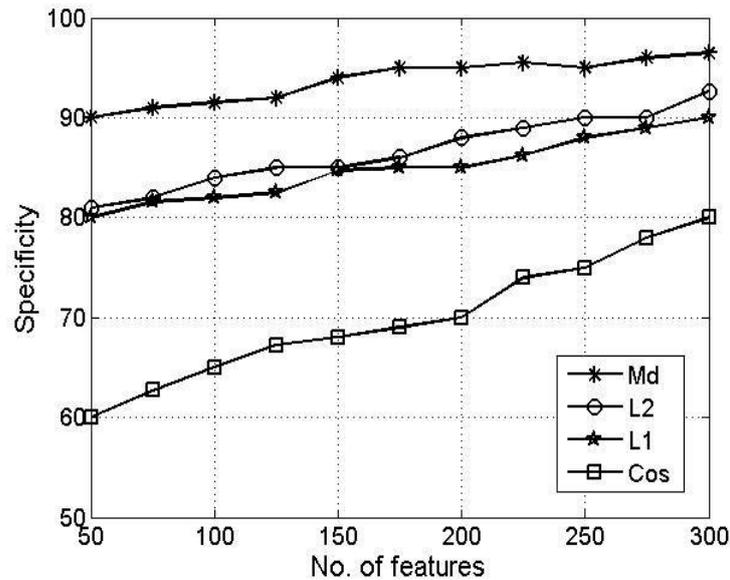

**Fig. 7.** Face recognition performance of the kernel ECA method with the cosine kernel function on the FERET database, using the four different similarity measures: $M_d$ (the Mahalanobis distance measure), $L_2$ (the $L_2$ distance measure), $L_1$ (the $L_1$ distance measure), and cos (the cosine similarity distance measure) for measuring negative recognition accuracy.

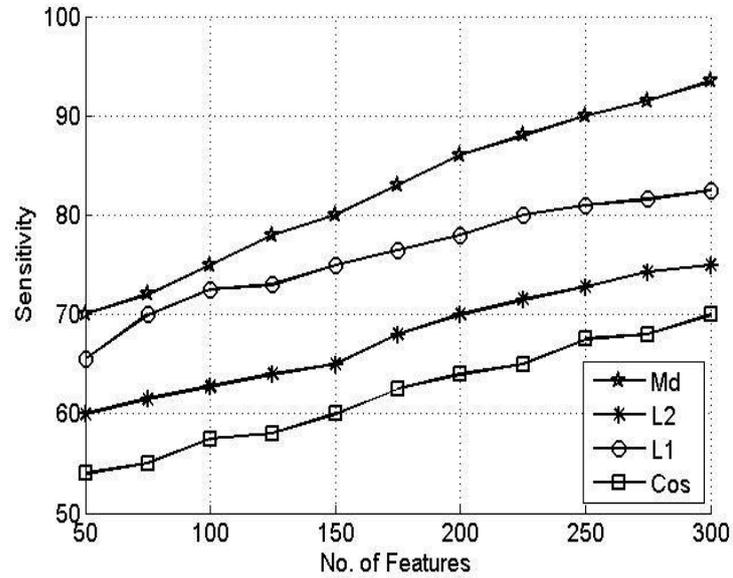

**Fig. 8.** Face recognition performance of the kernel ECA method with the cosine kernel function on the FERET database, using the four different similarity measures: $M_d$ (the Mahalanobis distance measure), $L_2$ (the $L_2$ distance measure), $L_1$ (the $L_1$distance measure), and cos (the cosine similarity measure) for measuring positive recognition accuracy.

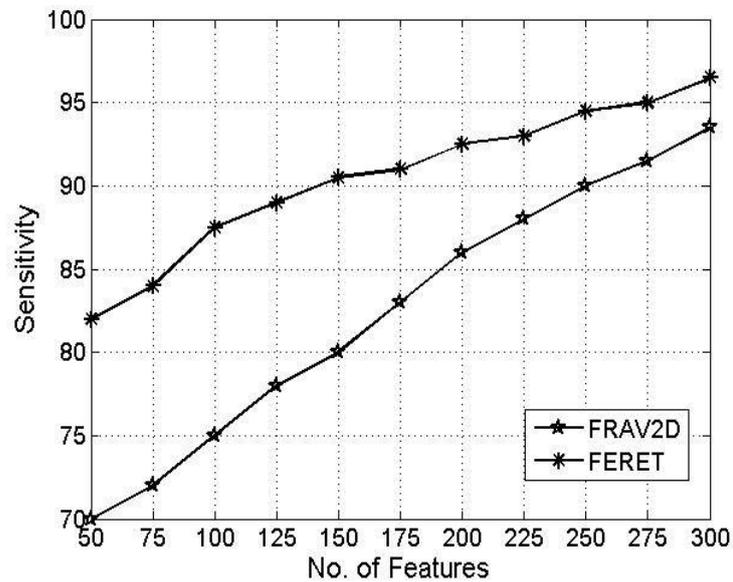

**Fig. 9**. Positive recognition performance of the proposed method using Mahalanobis distance measure on the dataset from the FRAV2D and FERET database.

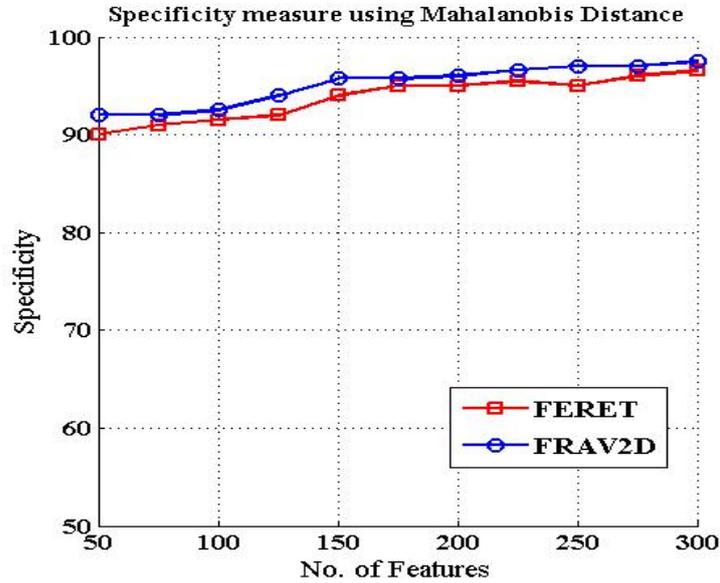

**Fig. 10**. Negative recognition performance of the proposed method using Mahalanobis distance measure on the dataset from the FRAV2D and FERET database.

The performance results of the proposed method, and the other tested algorithms on the FERET database as reported in [53] is presented in table 7.

**Table 7. Performance results of well-known algorithms and the Gabor based KECA method on the FERET database.**

| Method | Recognition Rate (%) |
|---|---|
| Eigenface method with Bayesian Similarity measure [53] | 79.0 |
| Elastic graph matching[53] | 84.0 |
| A Linear Discriminant Analysis based algorithm [53] | 88.0 |
| Kernel PCA[53] | 91.0 |
| Line based[53] | 92.7 |
| Neural network[53] | 93.5 |
| **Proposed Method** | **94.5** |

The images considered here, consists of frontal faces with different facial expressions, illumination conditions, and occlusions. Experimental results indicate that a) the extracted desirable discriminating Gabor features extracted by the proposed method enhances the face recognition performance in presence of occlusions as well as reduce computational complexity compared to EGM and other well known algorithms [53,54]. b) The KECA applied on the extracted feature vector further enhances recognition performance as KECA extracts only those features that preserve the maximum entropy.

# 5    Conclusion

This paper introduces a novel GWT based KECA method with cosine kernel function for both the frontal and pose-angled face recognition. Firstly GWT is applied on the image, as the GWT image has desirable facial features characterized by spatial frequency, spatial locality, and orientation selectivity to cope with the variations due to illumination and facial expression changes. From these GWT image features having the most important discriminating features are extracted. Then the kernel ECA extended with cosine kernel is applied on these most discriminating facial features to obtain the kernel ECA features which preserves the maximum entropy and hence further enhance face recognition. Further these extracted most discriminative facial features are compared locally instead of a general structure, hence it allows us to make a decision from the different parts of a face, and thus maximizes the benefit of applying the idea of "recognition by parts". Thus it performs well in presence of occlusions (such as sunglass, scarf etc.), the algorithm compares face in terms of mouth, nose and other features rather than the obstacle areas. This algorithm also reduces computational cost as KECA is applied, only on the extracted discriminating Gabor feature vectors of the GWT image rather than the whole image.

**Acknowledgement.** Authors are thankful to a major project entitled "Design and Development of Facial Thermogram Technology for Biometric Security System," funded by University Grants Commission (UGC),India and "DST-PURSE Programme" and CMATER and SRUVM project at Department of Computer Science and Engineering, Jadavpur University, India for providing necessary infrastructure to conduct experiments relating to this work.